\def\year{2017}\relax
\documentclass[letterpaper]{article}
\pdfoutput=1
\usepackage{aaai17}
\usepackage{times}
\usepackage{helvet}
\usepackage{courier}
\usepackage{url}  
\usepackage{graphicx}  

\usepackage{fancyhdr}

\usepackage[noend]{algorithmic}
\newcommand{\LINEIF}[2]{%
    \STATE\algorithmicif\ {#1}\ \algorithmicthen\ {#2}%
}

\newcommand{\LINEELSE}[1]{%
    \STATE\algorithmicelse\ {#1}\
}

\begin{document}

%%% JUST FOR PERSONAL VERSION
\thispagestyle{fancy}
\fancyhf{}
\chead{Proceedings of the Tenth International Symposium on Combinatorial Search, pages 2-10. AAAI Press, June 2017.}
%%% JUST FOR PERSONAL VERSION

\title{Variable Annealing Length and Parallelism in Simulated Annealing}
\author{Vincent A. Cicirello\\
Computer Science and Information Systems\\
School of Business, Stockton University\\
101 Vera King Farris Drive, 
Galloway, NJ 08205\\
\url{http://www.cicirello.org/}
}

\maketitle
\begin{abstract}
In this paper, we propose: (a) a restart schedule for an adaptive simulated annealer, and (b) 
parallel simulated annealing, with an adaptive and parameter-free annealing schedule.  
The foundation of our approach is the Modified Lam annealing schedule, which adaptively controls the temperature parameter 
to track a theoretically ideal rate of acceptance of neighboring states.  A sequential implementation of Modified Lam simulated 
annealing is almost parameter-free.  However, it requires prior knowledge of the annealing length.  We eliminate this  
parameter using restarts, with an exponentially increasing schedule of annealing lengths.  We then extend this restart schedule to
parallel implementation, executing several Modified Lam simulated annealers in 
parallel, with varying initial annealing lengths, and our proposed parallel annealing length schedule.
To validate our approach, we conduct experiments on an 
NP-Hard scheduling problem with sequence-dependent setup constraints.  We compare our approach to fixed 
length restarts, both sequentially and in parallel.  
Our results show that our approach can achieve substantial performance gains, throughout the course of the run, 
demonstrating our approach to be an effective anytime algorithm.
\end{abstract}

\section{Introduction}

For many scheduling and optimization problems, metaheuristics such as simulated annealing (SA), genetic algorithms (GA),
ant colony optimization (ACO), tabu search, etc, offer a means of trading off guarantees of optimality in favor of efficiently finding
high quality solutions.  Such algorithms often exhibit anytime behavior, providing increasingly better solutions with increases in available time.

Metaheuristic behavior is usually controlled by several parameters.  GAs have mutation and crossover rates, among other parameters.  SA
has parameters that control the annealing schedule.  
ACO has parameters that balance the relative influence of heuristic guidance and the learned pheromone trails.
Control parameters can either be tuned beforehand by some process (automated or otherwise) or adapted dynamically using search feedback.
For example, there exist parameter control approaches for GA and other forms of evolutionary computation~\cite{EIBEN:EC-1999,Wessing2011,Aleti2013}, 
adaptive annealing schedules for SA~\cite{Lam99dac,Swartz93Thesis,BOYAN-THESIS}, among others.

Parallel implementation of many metaheuristics is straightforward, such as population-based algorithms like GA
and ACO, whose motivations come from naturally occurring distributed behavior.  While for others, parallel implementation is less obvious.
In this paper, we propose a new approach to parallel SA.  We execute several independent runs, with restarts, in parallel of an adaptive SA using the Modified
Lam~\cite{Swartz93Thesis,BOYAN-THESIS} annealing schedule.  The Modified Lam annealing schedule is nearly parameter-free, requiring only knowledge of the annealing length.
In our proposed parallel SA, we eliminate this last parameter with restarts following a schedule of annealing lengths that better balance the risk associated with errors in
estimating available computation time.  The initial and restart annealing lengths increase exponentially.
The result is a parallel parameter-free SA with improved anytime behavior.

We validate our approach using an NP-Hard scheduling problem with sequence-dependent setups.  We begin our experiments with 
the sequential case.  Although for any a priori known fixed time limit, a single run of that length outperforms our 
restart schedule at run end, our proposed restart schedule exhibits greatly improved anytime behavior during the run.  We continue our experiments 
in parallel, demonstrating our parallel variable-length SA to significantly dominate parallel fixed-length restarts early in the run. 

The paper is organized as follows.  In Section~\ref{sec:related}, we discuss related work on parallel SA.  We provide details of our
parallel parameter-free SA in Section~\ref{sec:approach}.  Then, in Section~\ref{sec:exp}, we validate our approach experimentally with 
an NP-Hard scheduling problem, consisting of sequence-dependent setup constraints, and offer conclusions in Section~\ref{sec:conclude}.

\section{Background}\label{sec:related}

SA is typically described in a very sequential manner, and not parallelized as obviously as other metaheuristics.
There are two major categories of parallel SA~\cite{Rudolph1993}: namely, 
approaches that implement neighbor evaluations in parallel, and approaches that are essentially parallel implementations of multistart SA.
An example of the first case is the work of Ludwin and Betz who use a parallel SA for FPGA placement to optimize critical path length~\cite{Ludwin2011}.
They execute multiple move evaluations in parallel using what they call speculative moves.

The second type can be referred to as parallel multistart.
Although restarting some algorithms, such as hill climbers, offers a means of countering large numbers of local optima; 
many have shown sequential multistart SA to be ineffective.  
One long run of SA the length of the available time is typically more effective than taking the best solution from a set of shorter runs.
Therefore, it is not surprising that approaches to parallel SA that execute multiple runs of SA in parallel rarely
involve independent runs.  More commonly, parallel multistart SA involves sharing information among the parallel runs.
For example, Ram et al's approach (for job shop scheduling) periodically exchanges the best solution among the parallel runs, 
each continuing its search from there~\cite{Ram1996}.
More recently, in Jha and Menon's approach, at intervals called ``beats''
the best solution is shared among threads~\cite{Jha2014}.  Jha and Menon developed their approach for general purpose computation 
on graphics processing units (GPGPU) for a sports league scheduling problem.

Others apply SA in parallel to optimize a set of sub-problems, each SA instance solving a different sub-problem.  For example,
Rahimian et al's approach to graph partitioning, specifically for large social network graphs, distributes the problem, and individual distributed instances of
SA optimize portions of the problem~\cite{Rahimian2015}.  They apply this to both edge-cut and vertex-cut partitioning.

Other forms of search often rely on restarts, quite effectively.  For example, in constraint satisfaction, satisfiability, and other similar problems,
backtracking search using randomized variable-ordering and value-ordering heuristics often exhibit heavy-tailed runtime distributions~\cite{Gomes2000}.
Using an effective restart strategy, one can try to abandon the runs that are likely in the heavy-tail, restarting in an attempt to more
directly solve the problem.  The Luby restart schedule is the most widely known~\cite{Luby1993}, and has been parallelized~\cite{Cire2014}. 
The first several restart lengths of the Luby schedule (in number of backtracks) are as follows: $[ 1, 1, 2, 1, 1, 2, 4, 1, 1, 2, 1, 1, 2, 4, 8, \ldots ]$.
You begin with a restart sequence, $[ 1, 1, 2 ]$, then repeat the entire prior sequence from the beginning followed by a restart double the length of the last,
etc.

\section{Technical Approach}\label{sec:approach}

\subsection{Modified Lam Simulated Annealing}

The foundation of our approach is an existing sequential
simulated annealer with an adaptive annealing schedule.  SA operates via a mechanism
modeled after the process of heating a metal and allowing it to cool slowly.  
Heating enables the material to be shaped as desired, while cooling at a slow rate minimizes internal 
stress thus enabling greater stability in the final state.  

In SA, search is controlled by a temperature parameter.  The most basic form of SA begins with a high temperature
and then ``cools'' at some rate, with both initial temperature and cooling rate as system parameters.  The Modified
Lam annealing schedule~\cite{Swartz93Thesis,BOYAN-THESIS} eliminates these parameters, by dynamically adjusting 
temperature using search feedback.  It's based on results of Lam and Delosme~\cite{Lam99dac},
where they showed that the ideal run of SA accepts neighboring states at a rate of 0.44, which formed the basis for
an annealing schedule that tracks this acceptance rate.  Lam and Delosme's version originally used 
a monotonically decreasing temperature schedule, and adjusted the size of the neighborhood to maintain
the acceptance rate as near 0.44 as possible---e.g., they increased the size of the local
neighborhood to decrease the acceptance rate, and decreased the size of the local neighborhood
to increase the acceptance rate.  They relied on the common assumption
that nearby search states are of similar quality; and thus, a smaller local neighborhood
implies smaller difference between current fitness and neighbor fitness, which leads
to higher probability of neighbor acceptance.

Swartz later made additional observations on Lam and Delosme's annealing schedule~\cite{Swartz93Thesis},
which were then refined by Boyan into the Modified Lam schedule~\cite{BOYAN-THESIS}.
Specifically, Swartz observed that at the beginning of the search, the acceptance rate is near 1.0
(i.e., random search) and decreases at an exponential rate during the first 15\%
of the run when it reaches the target acceptance rate of 0.44, continues at that rate for 50\% of the run,
and then declines exponentially to the end of the run (i.e., end of run converges to a stochastic hill climb).
Rather than adjusting the size of the local neighborhood,
Swartz's and Boyan's Modified Lam schedule varies the temperature---increasing temperature to increase acceptance rate,
and decreasing temperature to decrease acceptance rate.  Figure~\ref{fig:lam} shows SA with the Modified Lam schedule.
In the pseudocode, $\eta(S)$ refers to the set of neighboring states of $S$ (i.e., our neighborhood function).

\begin{figure}[t]
\textbf{Modified Lam Annealing}%
\begin{algorithmic}
\STATE $S \leftarrow \mbox{GenerateRandomInitialState}$
\STATE $T \leftarrow 0.5$
\STATE $\mbox{AcceptRate} \leftarrow 0.5$
\FOR{$i = 1$ \textbf{to} $\mbox{MaxEvals}$}
	\STATE $S^{\prime} \leftarrow \mbox{random selection from} \; \eta(S)$
	\IF{$\mbox{Cost}(S^{\prime}) \leq \mbox{Cost}(S)$ \OR $\mbox{Rand} \in [0, 1) < e^{(\mbox{Cost}(S) - \mbox{Cost}(S^{\prime})) / T}$}
		\STATE $S \leftarrow S^{\prime}$
		\STATE $\mbox{AcceptRate} \leftarrow \frac{1}{500} (499 \cdot \mbox{AcceptRate} + 1)$
	\ENDIF\LINEELSE{$\mbox{AcceptRate} \leftarrow \frac{1}{500} (499 \cdot \mbox{AcceptRate})$}
	\IF{$i / \mbox{MaxEvals} < 0.15$}
		\STATE $\mbox{LamRate} \leftarrow 0.44 + 0.56 \cdot 560^{-i / \mbox{MaxEvals} / 0.15}$
	\ELSIF{$0.15 \leq i / \mbox{MaxEvals} < 0.65$}
		\STATE $\mbox{LamRate} \leftarrow 0.44$
	\ELSIF{$0.65 \leq i / \mbox{MaxEvals}$}
		\STATE $\mbox{LamRate} \leftarrow 0.44 \cdot 440^{-(i / \mbox{MaxEvals} - 0.65) / 0.35}$
	\ENDIF
	\LINEIF{$\mbox{AcceptRate} > \mbox{LamRate}$}{$T \leftarrow 0.999 \cdot T$}
	\LINEELSE{$T \leftarrow T / 0.999$}
\ENDFOR
\RETURN Best solution found during run
\end{algorithmic}
\caption{SA with the Modified Lam annealing schedule.}
\label{fig:lam}
\end{figure}

\subsection{Parallel Variable Length Runs}\label{sec:pSA}

The Modified Lam annealing schedule is nearly parameter-free.  However, it requires
the annealing length, referred to as $\mbox{MaxEvals}$ in the pseudocode of Figure~\ref{fig:lam}.
It is not always practical to accurately predict the time available for search.
If the run is shorter than you anticipate, the search would have spent too much time randomly exploring,
and insufficient time exploiting observed high quality portions of the search space.
If the run is much longer than you expected, it will get stuck too early in a local optimum, failing to
effectively utilize the unexpected extra time.

For sequential SA, many have shown that one longer run of SA is typically much better than restarting
shorter runs.  Extending this to parallel SA with independent instances, one would expect better performance
if all parallel instances were single runs of the available time.  However, the available time may not be known,
and may be difficult to accurately predict.  Our approach attempts to balance the risk associated with
incorrectly estimating time available, using restarts with a schedule of increasing annealing lengths.

Additionally, we define our restart schedule to support both sequential and parallel implementations.
Specifically, we propose a parallel SA, that executes several SA instances with the Modified Lam
annealing schedule.  Each SA instance has a different initial value of $\mbox{MaxEvals}$.
As each SA instance completes its initial run, it restarts with a new longer run.  The SA instances
operate independently, sharing no data, and each restart begins anew with randomly generated initial
solutions.  The best solution found among all parallel runs is returned.  The approach is essentially
a parallel implementation of a multistart SA, but where the length of the restarts varies and increases.

\paragraph*{Variable Annealing Length (VAL):}
In the sequential case, our annealing length schedule, VAL, is as follows.  
Restart $r$ ($r=0$ is the initial run) is of length:
\begin{equation}
\mbox{MaxEvals}(r) = 1000 * 2^r .
\end{equation}
Thus, the multistart SA follows a sequence
of annealing lengths $\{ 1000, 2000, 4000, 8000, 16000, 32000, \ldots \}$.

\paragraph*{Parallel Variable Annealing Length Version 0 (P-VAL-0):}
Assume that we execute $N$ instances, $\{ \mbox{SA}_0, \mbox{SA}_1, \ldots, \mbox{SA}_{N-1} \}$, 
of multistart Modified Lam SA in parallel.
The length, $\mbox{MaxEvals}_i(r)$, for restart $r$ of instance $\mbox{SA}_i$ is:
\begin{equation}
\mbox{MaxEvals}_i(r) = 1000 * 2^{i+r*N} .
\end{equation}
In the case of $N=1$, there is a single instance $\mbox{SA}_0$ and thus P-VAL reduces to VAL.

Consider $N=3$ as an example.  $\mbox{SA}_0$ has a sequence of
annealing lengths $\{ 1000, 8000, 64000, \ldots \}$, $\mbox{SA}_1$ has annealing lengths 
$\{ 2000, 16000, 128000, \ldots \}$, and $\mbox{SA}_2$ has annealing lengths 
$\{ 4000, 32000, 256000, \ldots \}$.  

\paragraph*{Parallel Variable Annealing Length (P-VAL):}
There are deficiencies in P-VAL-0 related to parallel speedup, for $N>4$, which we discuss later in Section~\ref{sec:v0v1}.
We resolve those deficiencies with the following schedule of annealing lengths
$\mbox{MaxEvals}_i(r)$, for restart $r$ of instance $\mbox{SA}_i$:
\begin{equation}
\mbox{MaxEvals}_i(r) = 1000 * 2^{(i \bmod 4) + r*\min\{N,4\}} .
\end{equation}
When $N \leq 4$, P-VAL is identical to P-VAL-0.  When $N>4$, 
$\{ \mbox{SA}_0, \mbox{SA}_4, \mbox{SA}_8, \ldots \}$ all have a sequence of
annealing lengths $\{ 1000, 16000, 256000, \ldots \}$,
$\{ \mbox{SA}_1, \mbox{SA}_5, \mbox{SA}_9, \ldots \}$ have
annealing lengths $\{ 2000, 32000, 512000, \ldots \}$, 
$\{ \mbox{SA}_2, \mbox{SA}_6, \mbox{SA}_{10}, \ldots \}$ have
annealing lengths $\{ 4000, 64000, 1024000, \ldots \}$, and
$\{ \mbox{SA}_3, \mbox{SA}_7, \mbox{SA}_{11}, \ldots \}$ have
annealing lengths $\{ 8000, 128000, 2048000, \ldots \}$.

\section{Experiments}\label{sec:exp}

\subsection{Scheduling with Sequence-Dependent Setups}

To validate our approach, we consider an NP-Hard single machine scheduling problem, characterized by
sequence-dependent setups, with an objective of minimizing weighted tardiness.
The problem is NP-Hard even if setups are independent of job ordering~\cite{BOOK-MORTON-93}, 
and the sequence-dependent setups magnify computational difficulty by inducing a
non-order-preserving property of the evaluation function~\cite{SEN-AIJ96}.

The problem is defined as follows, and consists of  
$N$ jobs, $J = \{ j_1, j_2, \ldots, j_N \}$.
Each job $j_k$ has weight $w_k$, duedate $d_k$, and processing time $p_k$. 
Setup times $s_{i,k}$, 
required prior to processing job $j_k$ if it immediately follows job $j_i$,
depend on the job ordering, and  
are asymmetric (i.e., $s_{i,k} \neq s_{k,i}$); and 
$s_{0,k}$ is the initial setup time required if job $j_k$ is processed first.
The jobs $J$ must be sequenced to minimize:
\begin{equation}\label{eq:wt}
T = \sum_{k=1}^{N} w_k T_k = \sum_{k=1}^{N} w_k \max(c_k - d_k, 0) ,
\end{equation}
where $T_k$ is the tardiness of job $j_k$.
The completion time $c_k$ of $j_k$ is the sum of the processing and setup times of $j_k$ and of all
jobs that preceed $j_k$. 
If $\pi(k)$ is the position of job $j_k$ in the sequence, then define $c_k$ as:
\begin{equation}
c_k = \sum_{\pi(x) \leq \pi(k), \pi(x)=\pi(y)+1} (p_x + s_{y,x}) .
\end{equation}

In our experiments, we use the standard benchmark set for the problem~\cite{CMU-THESIS,TECH2003},   
which consists of 120 instances, 40 each of loose, medium, and tight due dates.  Of these, 22  
loose duedate instances have an optimal weighted tardiness of 0.

The best available exact solver, Tanaka and Araki's Successive Sublimation Dynamic Programming,
can solve all of the available benchmark instances, but requires over two weeks of memory-intensive 
CPU time to solve the hardest instances~\cite{Tanaka2013}.  Therefore, metaheuristics
are a more practical approach.
A variety of algorithms have been proposed for the problem, 
such as dynamic programming \cite{Tanaka2013}, neighborhood search \cite{Liao2012}, iterated local search \cite{Xu2014}, value-biased stochastic sampling \cite{VBSS},
ACO \cite{Liao2007}, GA \cite{BICT2015,GECCO2006}, SA \cite{Cicirello-CP-Workshop-2007}, etc. 

We preprocess the instances transforming them as suggested by others~\cite{Tanaka2013,BICT2015}. 
To minimize the impact of setup times,
we increase the processing time of each job, $j_k$, 
by its minimum setup time, and reduce all setup times accordingly~\cite{BICT2015}:
\begin{equation}
s_k^{\min} = \min_{0\leq i \leq N, i \neq k} s_{i,k} , 
\end{equation}
\begin{equation}
p_k = p_k + s_k^{\min} , 
\end{equation}
\begin{equation}
s_{i,k} = s_{i,k} - s_k^{\min}, \forall i, i\neq k, 0\leq i \leq N .
\end{equation} 
We also eliminate jobs $j_k$, such that $w_k=0$, 
if $\forall x \forall y, x \neq y, s_{x,k} + p_k + s_{k,y} \geq s_{x,y}$~\cite{Tanaka2013}.

\subsection{Experimental Design}

We conduct our experiments on an Ubuntu 14.04 Server, with 32GB memory and
two Intel Xeon L5520 Quad-Core CPUs (2.27GHz).  The L5520 supports hyper-threading with two threads per core,
so our server has a total of 16 logical cores.  We implement our experiments with Java 8 and the 
Java HotSpot 64-bit Server VM.

We conduct experiments in both the sequential case ($N=1$) as well as in parallel.
For the parallel runs, we consider both $N=4$ and $N=8$ parallel instances.
We record the best solution found at 1 second intervals over 60 seconds.
   
We compare our VAL and P-VAL to multistart SA with fixed annealing length (FAL and P-FAL in parallel).
We consider several fixed annealing lengths.
FAL-1 and P-FAL-1 use an annealing length tuned to the total available time (60 seconds), 
specifically runs of 108 million SA evaluations.
This threshold was determined based on the total number of evaluations that VAL was able to do in 60 seconds.
Likewise, FAL-1/2, FAL-1/4, FAL-1/8, use annealing lengths that are $1/2$, $1/4$, and $1/8$ of the available
60 second limit (54 million, 27 million, and 13.5 million SA evaluations, respectively), restarting at that same
length as long as time remains.  P-FAL-1, P-FAL-1/2, P-FAL-1/4, P-FAL-1/8
are equivalent to a best of $N$, $2N$, $4N$, and $8N$ independent runs,
respectively, with approximately the same total cost.

\begin{figure}[t]
\includegraphics{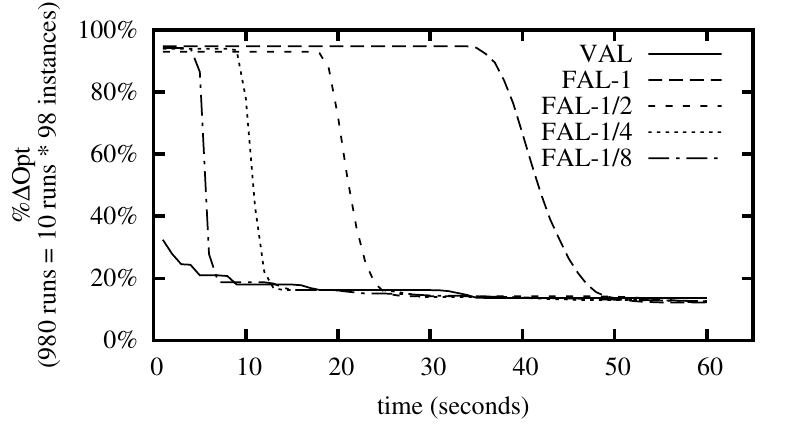}
\caption{Sequential case: $\%\Delta\mbox{Opt}$.}
\label{fig:opt1}
\end{figure}

\begin{figure}[t]
\includegraphics{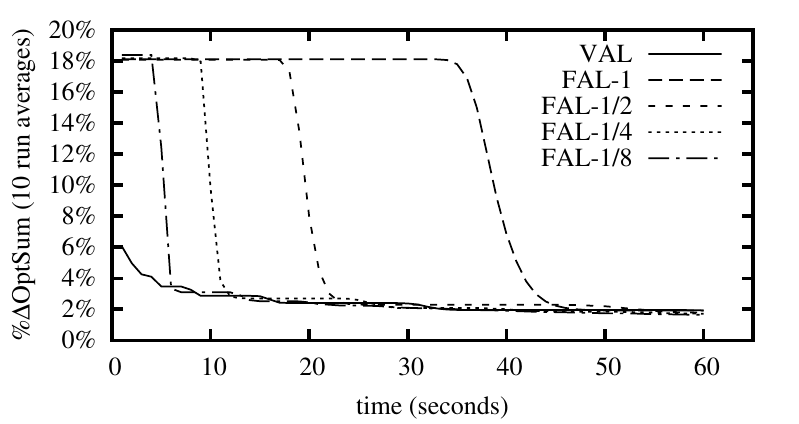}
\caption{Sequential case: $\%\Delta\mbox{OptSum}$.}
\label{fig:optsum1}
\end{figure}

We represent solutions as permutations, and use Insertion Mutation as our neighborhood function.  This operator
removes a random element, and reinserts it at a different random position.
Several existing metaheuristics for this scheduling problem use this operator, and it has been shown
effective for permutation optimization problems characterized by asymmetric edges and general position within
the permutation~\cite{TEVC2016}.  This problem has both: sequence-dependent setups (asymmetric edges), and due dates influence
general position within permutation.

We use the following commonly
employed metrics in the analysis of our experiments for this scheduling problem.  Most commonly reported is the average
percentage deviation from the optimal solutions, averaged only across the 98 instances with non-zero optimal values:
\begin{equation}
\%\Delta\mbox{Opt} = \frac{100}{N} \sum_{i=1}^{N} \frac{(S_i-O_i)}{O_i} ,
\end{equation}
where $S_i$ and $O_i$ are the value of the solution found for problem instance $i$ and its optimal solution, respectively.
One issue with this metric is that it ignores the 22 instances whose optimal solutions have weighted tardiness of 0. 
Thus, we also report the percentage deviation of the sum across all 120 instances relative to the sum of the optimal solutions:
\begin{equation}
\%\Delta\mbox{OptSum} = 100 \frac{\sum_{i=1}^{N} S_i - \sum_{i=1}^{N} O_i}{\sum_{i=1}^{N} O_i} .
\end{equation}

Using each algorithm, we optimize each instance 10 times.
We use t-tests to test the significance of the $\%\Delta\mbox{OptSum}$ results.  
We use the Wilcoxon signed rank test to test the significance of the $\%\Delta\mbox{Opt}$. Since $\%\Delta\mbox{Opt}$
is an average across multiple problem instances with values of varying scale, the t-test's normality requirement is not met.

\subsection{Sequential Results}

\begin{figure}[t]
\includegraphics{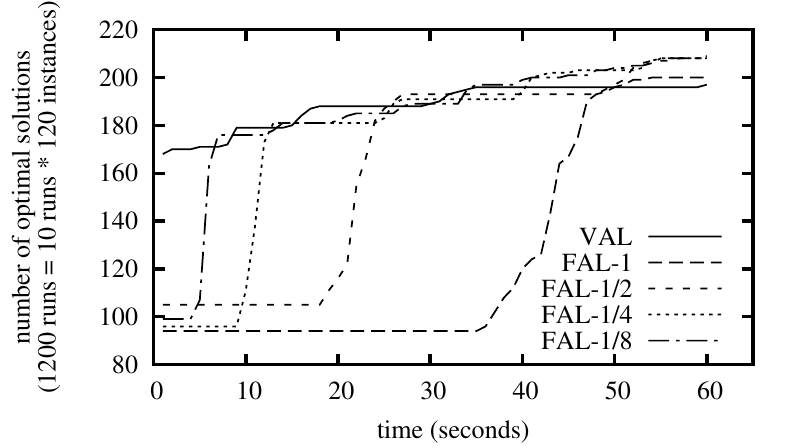}
\caption{Sequential case: Number of optimals.}
\label{fig:numopt1}
\end{figure}

The results for sequential SA are summarized in
Figures~\ref{fig:opt1} and~\ref{fig:optsum1}, which show average $\%\Delta\mbox{Opt}$ and $\%\Delta\mbox{OptSum}$, respectively,
throughout the duration of the 60 second runs. 
Early in the run, VAL dominates, 
and then performance approximately tracks that of each progressively longer fixed annealing length.  

On the $\%\Delta\mbox{Opt}$ metric, VAL dominates early in the run, but the FAL variations
overtake it at approximately the time corresponding to the annealing length for which they were tuned.
However, as VAL's longer runs complete, VAL's performance then matches that of the fixed length restarts.
For example, consider VAL versus FAL-1/8.  For the first 6 seconds, VAL strongly dominates at extremely significant levels
(p-values: $<10^{-18}$).  FAL-1/8 then outperforms VAL for approximately 2 seconds at significant levels (p-value at
second 8 is $<0.0001$).  From that point onward, there are periods where VAL's increasingly longer runs
enable it to gain a performance advantage over FAL-1/8 (at significant levels), and stretches with no significant performance difference.
Next, consider VAL versus FAL-1, fixed annealing length as long as the experiment.
FAL-1 catches up to VAL in performance (on $\%\Delta\mbox{Opt}$) at second 50, and its end of run performance is best
among all variations considered.  However, the end of run differences are not significant.  At one second intervals, 
from second 48 to the end of the run, p-values (from Wilcoxon signed rank test) between VAL and FAL-1 are no lower than $0.06$
and are as high as $0.86$.

The results on $\%\Delta\mbox{OptSum}$ are similar.  For example, for the first 48 seconds, VAL outperforms FAL-1 at significant levels
(t-test p-values from near zero early on to $p=0.018$ at second 48).  However, from second 49 to the end of the run, the difference in performance between
VAL and FAL-1 is not significant (p-values $>0.35$).  

Figure~\ref{fig:numopt1} shows the number of optimal solutions found out of 1200 runs as a function of time. 
VAL dominates early in the run, finding more optimal solutions than the others.  At the end of the run, the FAL variations
all find optimal solutions slightly more often than VAL.

In the sequential case, the restart schedule of annealing lengths enables SA to approximate the performance of long runs near the end of the run,
while simultaneously obtaining huge performance gains earlier in the run.

\begin{figure}[t]
\includegraphics{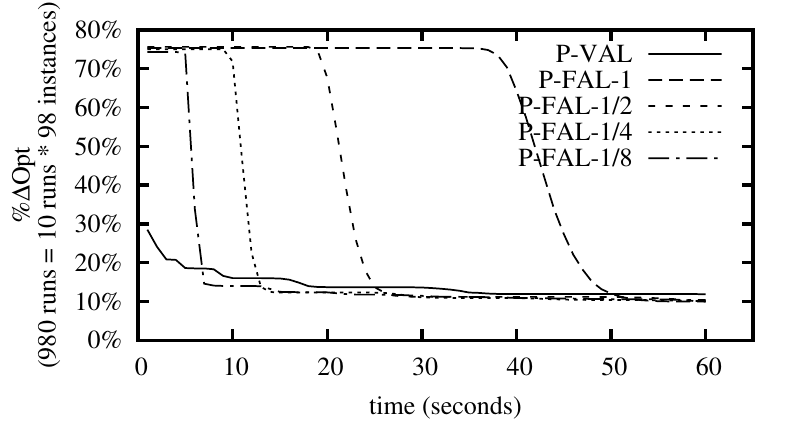}
\caption{Parallel case ($N=4$): $\%\Delta\mbox{Opt}$.}
\label{fig:opt4}
\end{figure}

\begin{figure}[t]
\includegraphics{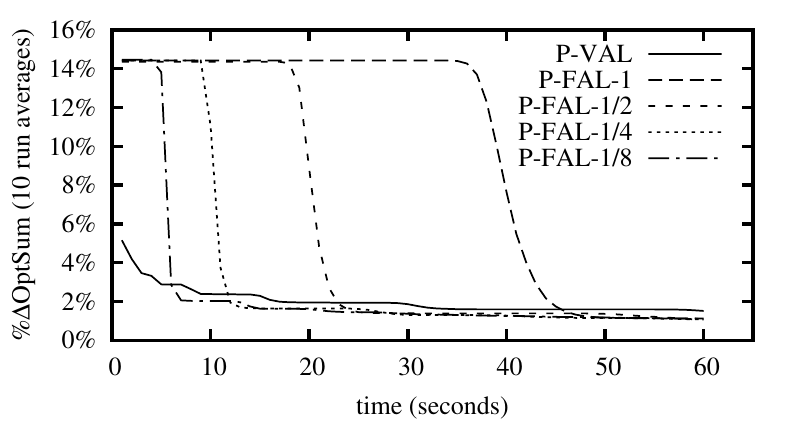}
\caption{Parallel case ($N=4$): $\%\Delta\mbox{OptSum}$.}
\label{fig:optsum4}
\end{figure}

\subsection{Parallel Results: 4 Parallel Instances}

Figures~\ref{fig:opt4} and~\ref{fig:optsum4} show average $\%\Delta\mbox{Opt}$ and $\%\Delta\mbox{OptSum}$, respectively,
for the duration of the 60 second runs for $N=4$ parallel instances.
Among the P-FAL variations, P-FAL-1 has the best performance on both metrics at the end of the run. However, the end of run differences
among the P-FAL variations are not statistically significant (p-values $> 0.13$ for $\%\Delta\mbox{Opt}$ and
p-values $> 0.08$ for $\%\Delta\mbox{OptSum}$).  Early in the run, P-VAL dominates relative to any fixed annealing length.
For 90\% of the run, P-VAL strongly dominates P-FAL-1, and dominates P-FAL-1/2 for nearly half the run, on both metrics, and similarly for
P-FAL-1/4 and P-FAL-1/8.
But late in the run, P-VAL does not match the performance
of fixed annealing length as well as it did in the sequential case. 
Once the fixed annealing length is reached, P-FAL outperforms to end of run, while P-VAL exhibits superior performance during the run.
All differences in performance between P-VAL and the P-FAL variations at each 1 second interval, 
except where the P-FAL curves cross the P-VAL curve, are statistically significant (p-values $< 0.0001$).

Although in the sequential case, VAL approximates the end of run performance of long fixed length runs, in parallel P-VAL
is outperformed at the end of the run by the long fixed length runs.  
P-VAL's advantage, however, is in improved anytime performance early in the run.

\subsection{P-VAL and P-VAL-0: 8 Parallel Instances}\label{sec:v0v1}

\begin{figure}[t]
\includegraphics{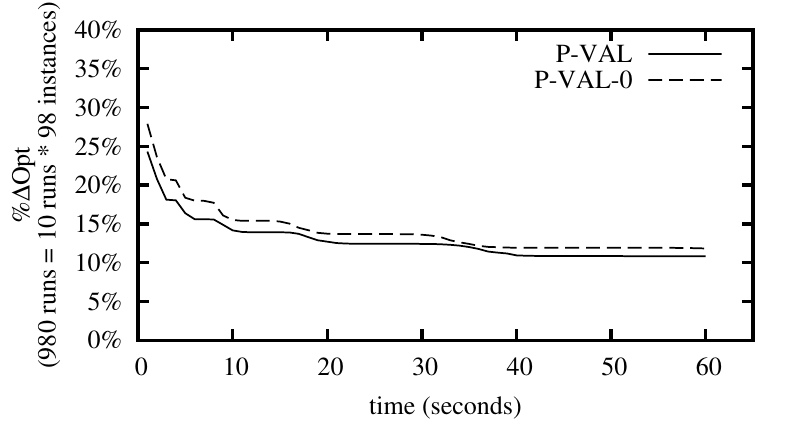}
\caption{Parallel case ($N=8$): $\%\Delta\mbox{Opt}$.}
\label{fig:optv0v1}
\end{figure}

\begin{figure}[t]
\includegraphics{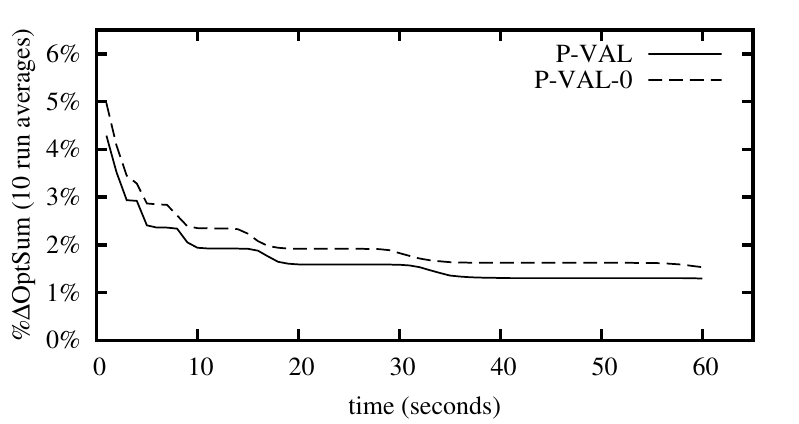}
\caption{Parallel case ($N=8$): $\%\Delta\mbox{OptSum}$.}
\label{fig:optsumv0v1}
\end{figure}

Earlier, we indicated that deficiencies exist in P-VAL-0 when $N>4$.  We consider this further here.
First, examine the performance of P-VAL-0 relative to P-VAL in the case of 8 parallel instances.
Figures~\ref{fig:optv0v1} and~\ref{fig:optsumv0v1} show average $\%\Delta\mbox{Opt}$ and $\%\Delta\mbox{OptSum}$, respectively,
for the duration of the 60 second runs.  
P-VAL strongly dominates P-VAL-0 throughout the run, on both metrics, at extremely statistically significant
levels (p-values $<0.00001$ at every one second interval).

Why is this the case?  For P-VAL-0, restart $r$ of $\mbox{SA}_i$ is of length 
$1000 * 2^{i+r*N}$, and completes
at time proportional to:
\begin{equation}
C_i(r) = 1000 \sum_{j=0}^{r} 2^{i+j*N} = 1000 * 2^i * \frac{2^{N(r+1)}-1}{2^N-1},
\end{equation}
which is the sum of the annealing lengths up to and including restart $r$.
The restart, $r_0$, of the sequential VAL that is of length $1000 * 2^{i+r*N}$
is $r_0 = i+r*N$, and completes
at time proportional to:
\begin{equation}
C_0(r_0) = 1000 \sum_{j=0}^{i+r*N} 2^j = 1000 * (2^{i+r*N+1} - 1) ,
\end{equation}
the sum of the run lengths up to and including restart $r_0$.
If parallel speedup was impacted only by completing longer runs sooner,
then the expected speedup factor is:
\begin{equation}
\frac{C_0(r_0)}{C_i(r)} = \frac{(2^{i+r*N+1} - 1)(2^N-1)}{2^i (2^{N(r+1)}-1)} .
\end{equation}
In the limit as the number of restarts $r$ grows large, the speedup due to completing
longer runs earlier is:
\begin{equation}
\lim_{r \rightarrow \infty} \frac{C_0(r_0)}{C_i(r)} =  \frac{2^N - 1}{2^{N-1}} .
\end{equation}
For $N=4$, the anticipated speedup from completing longer runs earlier
is 1.875, and for $N=8$ is approximately 1.992.  In fact, in the limit as the number of
parallel instances $N \rightarrow \infty$, the speedup factor approaches 2.0.  Specifically,
the longest completed restart finishes in half the time relative to the sequential VAL.
With $N=4$, P-VAL-0 is already approaching this limiting behavior, maximizing the benefit
from shortening the time for the longest runs to complete.

\begin{figure}[t]
\includegraphics{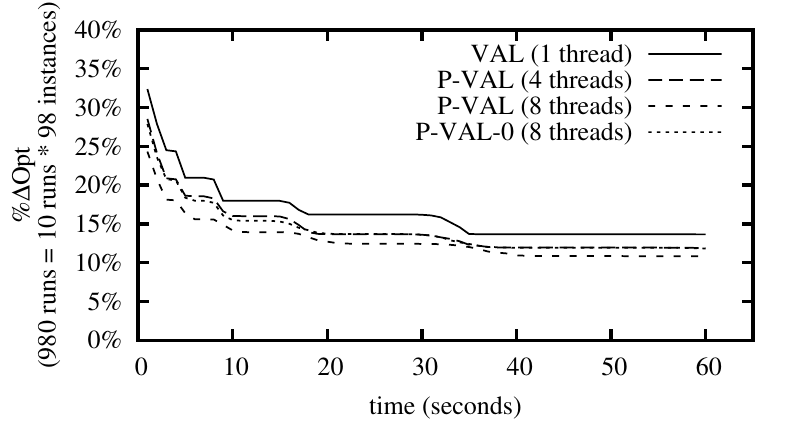}
\caption{VAL vs P-VAL vs P-VAL-0: $\%\Delta\mbox{Opt}$.}
\label{fig:opt148}
\end{figure}

\begin{figure}[t]
\includegraphics{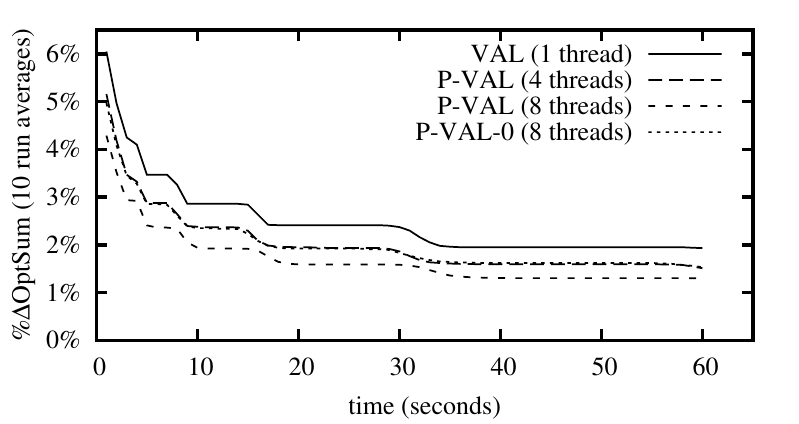}
\caption{VAL vs P-VAL vs P-VAL-0: $\%\Delta\mbox{OptSum}$.}
\label{fig:optsum148}
\end{figure}

We experimentally examine this in 
Figures~\ref{fig:opt148} and~\ref{fig:optsum148}, which show average $\%\Delta\mbox{Opt}$ and $\%\Delta\mbox{OptSum}$, respectively,
for a single sequential instance of VAL, P-VAL for both $N=4$ and $N=8$ parallel instances, as well as P-VAL-0 for $N=8$ (recall that
P-VAL and P-VAL-0 are identical for $N \leq 4$).
Visually, it is impossible to distinguish the P-VAL-0 results with $N=8$ from the $N=4$ case for both metrics.  The performance differences
between these two cases are not significant.  Using more than 4 parallel instances with P-VAL-0 does not improve performance.

The time for P-VAL with $N=4$ to reach a given $\%\Delta\mbox{Opt}$ (and likewise $\%\Delta\mbox{OptSum}$) is approximately one-half to
one-third the time taken by the sequential VAL (speedup factor between 2 and 3), rather than the one-fourth we would expect from linear speedup.  
This is consistent with the analysis above of completion time of the longest runs.
The speedup factor for P-VAL with $N=8$ is approximately 4 (the sequential VAL takes approximately 4 times as long to reach equivalent levels
of $\%\Delta\mbox{Opt}$ and $\%\Delta\mbox{OptSum}$).

As you can see, as we increase $N$ from 1 to 4 and then to 8, the difference in performance for P-VAL relative to VAL
is consistent throughout the run.  The speedup, through parallelization, is sublinear, but not limited by number of parallel instances.

\subsection{Parallel Results: 8 Parallel Instances}

\begin{figure}[t]
\includegraphics{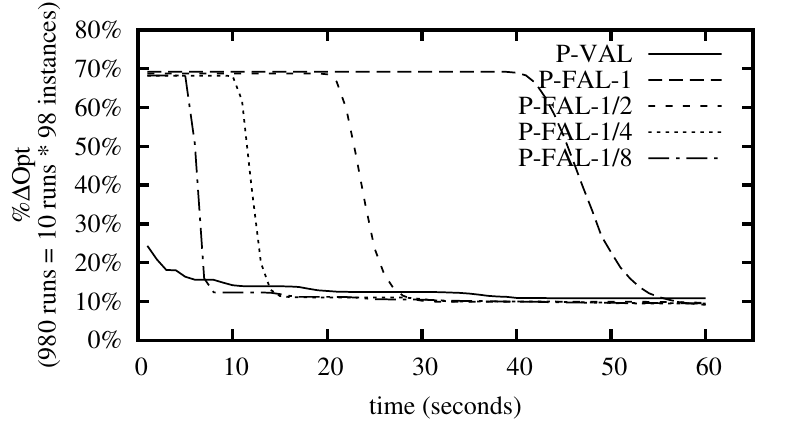}
\caption{Parallel case ($N=8$): $\%\Delta\mbox{Opt}$.}
\label{fig:opt}
\end{figure}

\begin{figure}[t]
\includegraphics{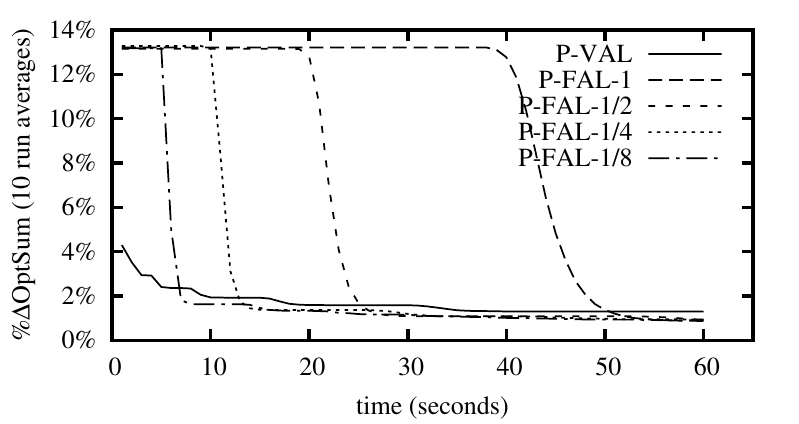}
\caption{Parallel case ($N=8$): $\%\Delta\mbox{OptSum}$.}
\label{fig:optsum}
\end{figure}

Now consider $N=8$ parallel instances.
Figures~\ref{fig:opt} and~\ref{fig:optsum} show average $\%\Delta\mbox{Opt}$ and $\%\Delta\mbox{OptSum}$, respectively,
for the duration of the 60 second runs.  For most of the run, P-VAL very strongly dominates.
Each of the P-FAL variations eventually overtake P-VAL as they near their tuned annealing lengths.
For example, P-FAL-1/2 overtakes P-VAL at second 26 on $\%\Delta\mbox{OptSum}$ and second 28 on $\%\Delta\mbox{Opt}$, and
P-FAL-1 overtakes P-VAL at second 51 on $\%\Delta\mbox{OptSum}$ and second 56 on $\%\Delta\mbox{Opt}$. 
Though not evident from the scale of the graphs, P-FAL-1 outperforms all others by end of run, at statistically significant levels,
on both metrics.
We should expect P-FAL-1 to perform best at the end, as P-FAL-1 executes 
multiple long runs in parallel, providing the Modified Lam schedule with the actual experimental run length. 
This is also consistent with other findings that show a long run of SA is usually better than multiple independent short runs.

However, for 90\% of the run, P-VAL strongly dominates P-FAL-1, and dominates P-FAL-1/2 for nearly half the run, on both metrics, and similarly for
P-FAL-1/4 and P-FAL-1/8.  P-VAL exhibits superior performance during the run.

\begin{figure}[t]
\includegraphics{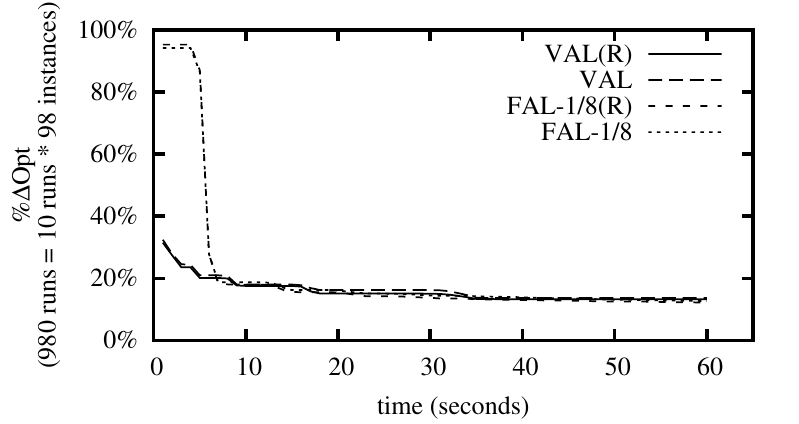}
\caption{Reannealing Sequential: $\%\Delta\mbox{Opt}$.}
\label{fig:opt1re}
\end{figure}

\begin{figure}[t]
\includegraphics{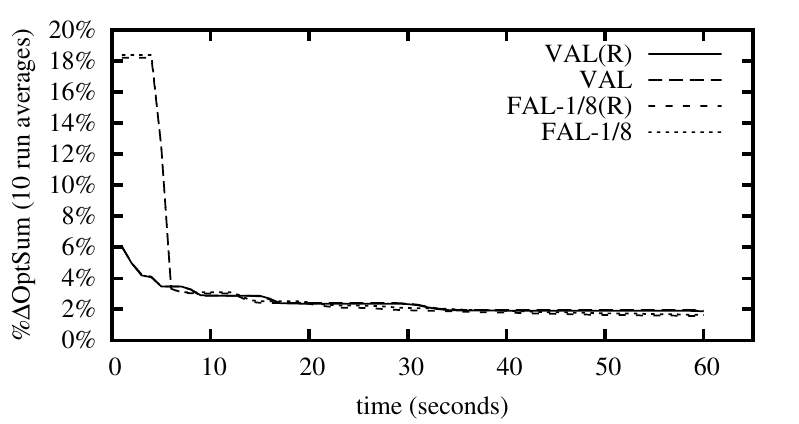}
\caption{Reannealing Sequential: $\%\Delta\mbox{OptSum}$.}
\label{fig:optsum1re}
\end{figure}

\begin{figure}[t]
\includegraphics{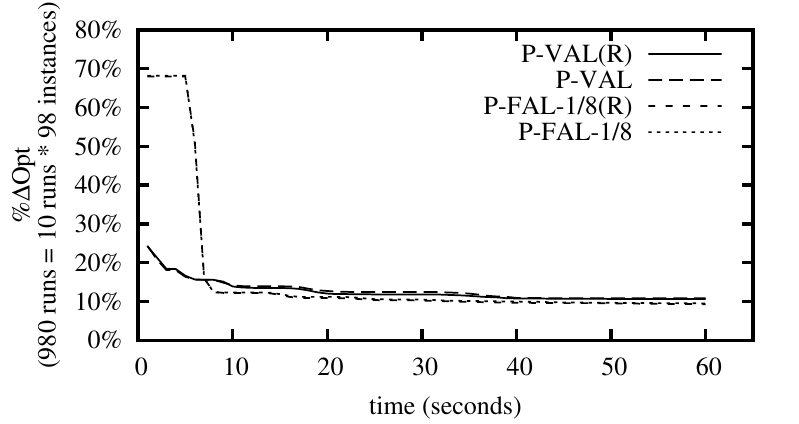}
\caption{Reannealing Parallel ($N=8$): $\%\Delta\mbox{Opt}$.}
\label{fig:opt8re}
\end{figure}

\begin{figure}[t]
\includegraphics{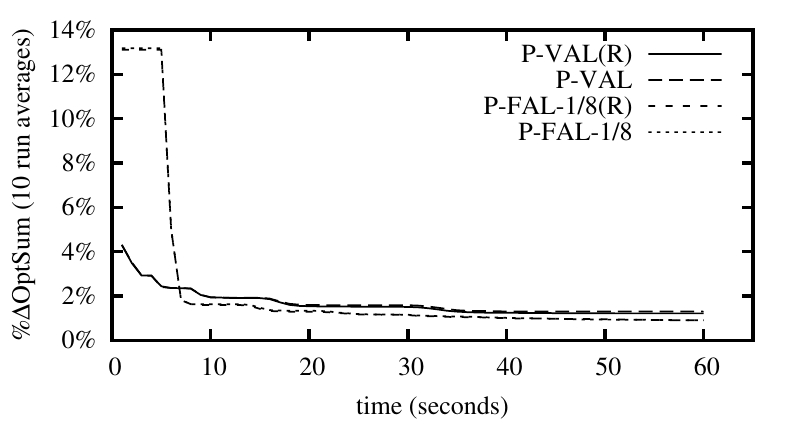}
\caption{Reannealing Parallel ($N=8$): $\%\Delta\mbox{OptSum}$.}
\label{fig:optsum8re}
\end{figure}

The $\%\Delta\mbox{OptSum}$ differences between P-VAL and P-FAL-1 are statistically significant (t-test p-values $<0.04$) at 
every one-second interval; and 
the $\%\Delta\mbox{Opt}$ results are significant (Wilcoxon signed rank test, p-values ranging from near-zero to 0.008), except where
the curves cross ($p=0.57$).  The differences between P-VAL and each of P-FAL-1/2, P-FAL-1/4, and P-FAL-1/8 are statistically significant at every 1 second 
interval.

\subsection{On the Efficacy of Reannealing}\label{sec:reanneal}

Thus far, all experimental results use independent restarts from random starting solutions, with no data
sharing among parallel instances.  We now explore what, if any, benefit is gained from reannealing
prior solutions.  Specifically, we consider VAL(R), where each restart reanneals the best of run solution 
rather than a random one.  P-VAL(R) reanneals the current best of run solution across all 
parallel instances.  Likewise, FAL-1/8(R) is FAL-1/8, but with reannealing of the best of run solution (similarly for P-FAL-1/8(R)).  

The sequential results are found in Figures~\ref{fig:opt1re} and~\ref{fig:optsum1re}, and the parallel results (for $N=8$)
are in Figures~\ref{fig:opt8re} and~\ref{fig:optsum8re}.
For variable annealing lengths, both sequential and parallel, 
the differences in performance between reannealing and random starting solutions 
are not statistically significant, except for the parallel case during the last 15-20 seconds of the run, where
reannealing leads to marginally better results at statistically significant levels (p-values less than 0.01).
Visually, in the graphs, it is virtually impossible to distinguish these.
For fixed annealing length, both sequential and parallel, the differences in performance with and without reannealing
are not statistically significant.  
Reannealing good solutions does not provide any benefit over independent runs in either the sequential or parallel case.

\section{Conclusions}\label{sec:conclude}

In this paper, we proposed a restart schedule for SA using the modified Lam annealing schedule.
Our restart schedule eliminates the need to know the annealing length a priori.  Relying on the often demonstrated
property of SA that single long runs typically outperform multiple short runs, our restart schedule increases the 
annealing length at an exponential rate.  The shorter runs at the beginning enable quickly
finding ``good'' solutions, while the increasing annealing lengths of the restarts enable approximating
the end of run performance of a single long run of SA. 

Our restart schedule supports parallel implementation, using parallel independent SA instances that vary in initial annealing length, 
and with exponentially increasing restart lengths.  The initial annealing lengths are staggered to ensure that longer runs are 
already in progress as shorter runs complete.  Our aim is to balance the risk associated with errors in determining the
time available for problem solving.

The end-of-run behavior observed in experiments, both sequential as well as in parallel, with a sequence-dependent scheduling problem 
confirm the commonly found property that a longer SA run outperforms restarts of a shorter run.
However, performance during the run is often overlooked.  For example, although FAL-1 performs best at the end
of run, FAL-1/2 achieved better results at the mid-way point.  Our annealing length schedule VAL, and in parallel P-VAL, exhibited
stronger anytime behavior throughout the run.

\bibliographystyle{aaai}

\begin{thebibliography}{}

\bibitem[\protect\citeauthoryear{Aleti and Moser}{2013}]{Aleti2013}
Aleti, A., and Moser, I.
\newblock 2013.
\newblock Entropy-based adaptive range parameter control for evolutionary
  algorithms.
\newblock In {\em Proc. GECCO},  1501--1508.
\newblock ACM.

\bibitem[\protect\citeauthoryear{Boyan}{1998}]{BOYAN-THESIS}
Boyan, J.~A.
\newblock 1998.
\newblock {\em Learning Evaluation Functions for Global Optimization}.
\newblock Ph.D. Dissertation, Carnegie Mellon University, Pittsburgh, PA.

\bibitem[\protect\citeauthoryear{Cicirello and Smith}{2005}]{VBSS}
Cicirello, V.~A., and Smith, S.~F.
\newblock 2005.
\newblock Enhancing stochastic search performance by value-biased randomization
  of heuristics.
\newblock {\em Journal of Heuristics} 11(1):5--34.

\bibitem[\protect\citeauthoryear{Cicirello}{2003a}]{CMU-THESIS}
Cicirello, V.~A.
\newblock 2003a.
\newblock {\em Boosting Stochastic Problem Solvers Through Online Self-Analysis
  of Performance}.
\newblock Ph.D. Dissertation, Carnegie Mellon University.

\bibitem[\protect\citeauthoryear{Cicirello}{2003b}]{TECH2003}
Cicirello, V.~A.
\newblock 2003b.
\newblock Weighted tardiness scheduling with sequence-dependent setups: A
  benchmark library.
\newblock Tech. report, ICL Lab, CMU.
\newblock \url{http://www.cicirello.org/datasets/wtsds/}.

\bibitem[\protect\citeauthoryear{Cicirello}{2006}]{GECCO2006}
Cicirello, V.~A.
\newblock 2006.
\newblock Non-wrapping order crossover: An order preserving crossover operator
  that respects absolute position.
\newblock In {\em Proc. GECCO'06}. ACM.
\newblock  1125--1131.

\bibitem[\protect\citeauthoryear{Cicirello}{2007}]{Cicirello-CP-Workshop-2007}
Cicirello, V.~A.
\newblock 2007.
\newblock On the design of an adaptive simulated annealing algorithm.
\newblock In {\em Proc. CP 2007 First Workshop on Autonomous Search}. AAAI
  Press.

\bibitem[\protect\citeauthoryear{Cicirello}{2015}]{BICT2015}
Cicirello, V.~A.
\newblock 2015.
\newblock Genetic algorithm parameter control: Application to scheduling with
  sequence-dependent setups.
\newblock In {\em Proc. 9th Int. Conf. Bio-inspired Information and
  Communications Technologies}. ICST.
\newblock  136--143.

\bibitem[\protect\citeauthoryear{Cicirello}{2016}]{TEVC2016}
Cicirello, V.~A.
\newblock 2016.
\newblock The permutation in a haystack problem and the calculus of search
  landscapes.
\newblock {\em IEEE Transactions on Evolutionary Computation} 20(3):434--446.

\bibitem[\protect\citeauthoryear{Cire, Kadioglu, and Sellmann}{2014}]{Cire2014}
Cire, A.~A.; Kadioglu, S.; and Sellmann, M.
\newblock 2014.
\newblock Parallel restarted search.
\newblock In {\em Proceedings of the Twenty-Eighth AAAI Conference on
  Artificial Intelligence},  842--848.
\newblock AAAI Press.

\bibitem[\protect\citeauthoryear{Eiben, Hinterding, and
  Michalewicz}{1999}]{EIBEN:EC-1999}
Eiben, A.~E.; Hinterding, R.; and Michalewicz, Z.
\newblock 1999.
\newblock Parameter control in evolutionary algorithms.
\newblock {\em IEEE Transactions on Evolutionary Computation} 3(2):124--141.

\bibitem[\protect\citeauthoryear{Gomes \bgroup et al\mbox.\egroup
  }{2000}]{Gomes2000}
Gomes, C.~P.; Selman, B.; Crato, N.; and Kautz, H.
\newblock 2000.
\newblock Heavy-tailed phenomena in satisfiability and constraint satisfaction
  problems.
\newblock {\em Journal of Automated Reasoning} 24(1):67--100.

\bibitem[\protect\citeauthoryear{Jha and Menon}{2014}]{Jha2014}
Jha, S., and Menon, V.
\newblock 2014.
\newblock Bbmttp: Beat-based parallel simulated annealing algorithm on gpgpus
  for the mirrored traveling tournament problem.
\newblock In {\em Proceedings of the High Performance Computing Symposium},
  3:1--3:7.
\newblock Society for Computer Simulation International.

\bibitem[\protect\citeauthoryear{Lam and Delosme}{1988}]{Lam99dac}
Lam, J., and Delosme, J.
\newblock 1988.
\newblock Performance of a new annealing schedule.
\newblock In {\em Proc. 25th ACM/IEEE DAC},  306--311.

\bibitem[\protect\citeauthoryear{Liao and Juan}{2007}]{Liao2007}
Liao, C.-J., and Juan, H.-C.
\newblock 2007.
\newblock An ant colony optimization for single-machine tardiness scheduling
  with sequence-dependent setups.
\newblock {\em Computers and Operations Research} 34(7):1899--1909.

\bibitem[\protect\citeauthoryear{Liao, Tsou, and Huang}{2012}]{Liao2012}
Liao, C.-J.; Tsou, H.-H.; and Huang, K.-L.
\newblock 2012.
\newblock Neighborhood search procedures for single machine tardiness
  scheduling with sequence-dependent setups.
\newblock {\em Theoretical Comp. Sci.} 434:45--52.

\bibitem[\protect\citeauthoryear{Luby, Sinclair, and
  Zuckerman}{1993}]{Luby1993}
Luby, M.; Sinclair, A.; and Zuckerman, D.
\newblock 1993.
\newblock Optimal speedup of las vegas algorithms.
\newblock {\em Information Processing Letters} 47(4):173--180.

\bibitem[\protect\citeauthoryear{Ludwin and Betz}{2011}]{Ludwin2011}
Ludwin, A., and Betz, V.
\newblock 2011.
\newblock Efficient and deterministic parallel placement for fpgas.
\newblock {\em ACM Trans. Des. Autom. Electron. Syst.} 16(3):22:1--22:23.

\bibitem[\protect\citeauthoryear{Morton and Pentico}{1993}]{BOOK-MORTON-93}
Morton, T.~E., and Pentico, D.~W.
\newblock 1993.
\newblock {\em Heuristic Scheduling Systems: With Applications to Production
  Systems and Project Management}.
\newblock Wiley.

\bibitem[\protect\citeauthoryear{Rahimian \bgroup et al\mbox.\egroup
  }{2015}]{Rahimian2015}
Rahimian, F.; Payberah, A.~H.; Girdzijauskas, S.; Jelasity, M.; and Haridi, S.
\newblock 2015.
\newblock A distributed algorithm for large-scale graph partitioning.
\newblock {\em ACM Trans. Auton. Adapt. Syst.} 10(2):12:1--12:24.

\bibitem[\protect\citeauthoryear{Ram, Sreenivas, and
  Subramaniam}{1996}]{Ram1996}
Ram, D.~J.; Sreenivas, T.~H.; and Subramaniam, K.~G.
\newblock 1996.
\newblock Parallel simulated annealing algorithms.
\newblock {\em Journal of Parallel and Distributed Computing} 37:207–212.

\bibitem[\protect\citeauthoryear{Rudolph}{1993}]{Rudolph1993}
Rudolph, G.
\newblock 1993.
\newblock Massively parallel simulated annealing and its relation to
  evolutionary algorithms.
\newblock {\em Evol. Comput.} 1(4):361--383.

\bibitem[\protect\citeauthoryear{Sen and Bagchi}{1996}]{SEN-AIJ96}
Sen, A.~K., and Bagchi, A.
\newblock 1996.
\newblock Graph search methods for non-order-preserving evaluation functions:
  Applications to job sequencing problems.
\newblock {\em AIJ} 86(1):43--73.

\bibitem[\protect\citeauthoryear{Swartz}{1993}]{Swartz93Thesis}
Swartz, W.~P.
\newblock 1993.
\newblock {\em Automatic Layout of Analog and Digital Mixed Macro/Standard Cell
  Integrated Circuits}.
\newblock Ph.D. Dissertation, Yale University.

\bibitem[\protect\citeauthoryear{Tanaka and Araki}{2013}]{Tanaka2013}
Tanaka, S., and Araki, M.
\newblock 2013.
\newblock An exact algorithm for the single-machine total weighted tardiness
  problem with sequence-dependent setup times.
\newblock {\em Computers and Operations Research} 40(1):344--352.

\bibitem[\protect\citeauthoryear{Wessing, Preuss, and
  Rudolph}{2011}]{Wessing2011}
Wessing, S.; Preuss, M.; and Rudolph, G.
\newblock 2011.
\newblock When parameter tuning actually is parameter control.
\newblock In {\em Proc. GECCO},  821--828.
\newblock ACM.

\bibitem[\protect\citeauthoryear{Xu, L\"{u}, and Cheng}{2014}]{Xu2014}
Xu, H.; L\"{u}, Z.; and Cheng, T.~C.
\newblock 2014.
\newblock Iterated local search for single-machine scheduling with
  sequence-dependent setup times to minimize total weighted tardiness.
\newblock {\em J. of Scheduling} 17(3):271--287.

\end{thebibliography}

\end{document}